\documentclass[10pt,journal,compsoc]{IEEEtran}
\usepackage{setspace}
\usepackage[utf8]{inputenc}
\usepackage[cmex10]{amsmath}
\usepackage{graphicx}
\usepackage{amssymb}
\usepackage{amsthm}
\usepackage{cite}
\usepackage{color}
\usepackage{caption}
\usepackage{amsmath}
\usepackage{tabu}
\usepackage{array}
\usepackage{booktabs}
\usepackage{amssymb}
\usepackage{threeparttable}
\usepackage[linesnumbered,ruled]{algorithm2e}
\usepackage{float}
\usepackage{stfloats}
\usepackage{subcaption}
\usepackage{lipsum}
\usepackage[english]{babel}
\usepackage{epstopdf}

\let\oldnl\nl% Store \nl in \oldnl
\newcommand{\nonl}{\renewcommand{\nl}{\let\nl\oldnl}}% Remove line number for one line

\newtheorem{thm}{Lemma}%here second one is the name you wanna use
\newtheorem{problem}{Problem}

\theoremstyle{definition}
\newtheorem{defn}{Definition}
\newtheorem{exmp}{Example} % same for example numbers

\begin{document}
\title{Privacy Preserving Location Data Publishing: A~Machine Learning Approach}
%\author{
%    \IEEEauthorblockN{Sina Shaham, Ming Ding, Bo Liu, Zihuai Lin, Jun Li}\\
%    \IEEEauthorblockA{
%        School of Electrical and Information Engineering, The University of Sydney, Australia \\
%        Department of Engineering, La Trobe University, Australia\\
%        Email: \{sina.shaham, zihuai.lin\}@sydney.edu.au, ming.ding@data61.csiro.au, b.liu2@latrobe.edu.au, jun.li@njust.edu.cn}

\author{Sina Shaham, Ming Ding, Bo Liu, Shuping Dang, Zihuai Lin, and Jun Li% <-this % stops a space
\thanks{This work was submitted in part and accepted to appear in the proceedings of INFOCOM WORKSHOPS, 2019 \cite{shaham2019machine}.}
\thanks{S. Shaham and Z. Lin are with the Department
of Engineering, The University of Sydney, Sydney, NSW, 2006 Australia (e-mail: sina.shaham, zihuai.lin\}@sydney.edu.au).}% <-this % stops a space
\thanks{M. Ding is with Data61, Sydney, NSW, 1435 Australia (email: ming.ding@data61.csiro.au)}
\thanks{B. Liu is with Latrobe University, VIC, 3086 Australia (email: B.Liu2@latrobe.edu.au)}
\thanks{S. Dang was with the R\&D Center, Guangxi Huanan Communication Co., Ltd., Nanning 530007, China when completing the major work of this paper, and is now with Computer, Electrical and Mathematical Science and Engineering Division, King Abdullah University of Science and Technology (KAUST), Thuwal 23955-6900, Kingdom of Saudi Arabia (e-mail: shuping.dang@kaust.edu.sa).}
\thanks{J. Li is with NJUST, Nanjing, China (email: jleesr80@gmail.com)}}

%        Email: \{matthew.kokshoorn, he.chen, yonghui.li,  branka.vucetic\}@sydney.edu.au }
%    \thanks{This research was supported by ARC grants DP150104019 and FT120100487. The research was also supported by funding from the Faculty of Engineering and Information Technologies, The University of Sydney, under the Faculty Research Cluster ProgRAF and the Faculty Early Career Researcher Scheme.}

\IEEEtitleabstractindextext{
\begin{abstract}
Publishing datasets plays an essential role in open data research and promoting transparency of government agencies. However, such data publication might reveal users' private information. One of the most sensitive sources of data is spatiotemporal trajectory datasets. Unfortunately, merely removing unique identifiers cannot preserve the privacy of users. Adversaries may know parts of the trajectories or be able to link the published dataset to other sources for the purpose of user identification. Therefore, it is crucial to apply privacy preserving techniques before the publication of spatiotemporal trajectory datasets. In this paper, we propose a robust framework for the anonymization of spatiotemporal trajectory datasets termed as machine learning based anonymization (MLA). By introducing a new formulation of the problem, we are able to apply machine learning algorithms for clustering the trajectories and propose to use $k$-means algorithm for this purpose. A variation of  $k$-means algorithm is also proposed to preserve the privacy in overly sensitive datasets. Moreover, we improve the alignment process by considering multiple sequence alignment as part of the MLA. The framework and all the proposed algorithms are applied to TDrive and Geolife location datasets. The experimental results indicate a significantly higher utility of datasets by anonymization based on MLA framework.
\end{abstract}
}
%}
\maketitle

\IEEEraisesectionheading{\section{Introduction}\label{Introduction}}
%\section{Introduction}\label{Introduction}

\IEEEPARstart{P}{ublication} of data by different organizations and institutes is crucial for open research and transparency of government agencies. Just in Australia, since 2013, over 7000 additional datasets have been published on 'data.gov.au,' a dedicated website for the publication of datasets by the Australian government. Moreover, the new Australian government data sharing legislation encourage government agencies to publish their data, and as early as 2019, many of them will have to do so \cite{aa}. Unfortunately, the process of data publication can be highly risky as it may disclose individuals' sensitive information. Hence, an essential step before publishing datasets is to remove any uniquely identifiable information from them. However, such an operation is not sufficient for preserving the privacy of users. Adversaries can re-identify individuals in datasets based on common attributes called quasi-identifiers or may have prior knowledge about the trajectories traveled by the users. Such side information enables them to reveal sensitive information that can cause physical, financial, and reputational harms to people.

One of the most sensitive sources of data is location trajectories or spatiotemporal trajectories. Despite numerous use cases that the publication of spatiotemporal data can provide to users and researchers, it poses a significant threat to users' privacy. As an example, consider a person who has been using GPS navigation to travel from home to work every morning of weekdays. If an adversary has some prior knowledge about a user, such as the home address, it is possible to identify the user. Such an inference attack can compromise user privacy, such as revealing the user's health condition and how often the user visits his/her medical specialist. Therefore, it is crucial to anonymize spatiotemporal datasets before publishing them to the public. The privacy issue gets even more severe if the adversary links identified users to other databases, such as the database of medical records. That is the very reason why nowadays most companies are reluctant to publish any spatiotemporal trajectory datasets without applying an effective privacy preserving technique.

A widely accepted privacy metric for the publication of spatiotemporal datasets is $k$-anonymity. This metric can be summarized as ensuring that every trajectory in the published dataset is indistinguishable from at least $k-1$ other trajectories. The authors in \cite{medical}, adopted the notion of $k$-anonymity for spatiotemporal datasets and proposed an anonymization algorithm based on generalization.  Xu et al. \cite{xu2017trajectory} investigated the effects of factors such as spatiotemporal resolution and the number of users released on the anonymization process. Dong et al.~\cite{dong2018novel} focused on improving the existing clustering approaches. They proposed an anonymization scheme based on achieving $k$-anonymity by grouping similar trajectories and removing the highly dissimilar ones. More recently, the authors in \cite{comparison} developed an algorithm called k-merge to anonymize the trajectory datasets while preserving the privacy of users from probabilistic attacks. Local suppression and splitting techniques were also considered to protect privacy in \cite{terrovitis2017local}.

However, there are three major problems with the aforementioned approaches.
\begin{itemize}
  \item Lack of a well-defined method to cluster trajectories as there is not an easy way to measure the cost of clustering when considering the distances among trajectories rather than simply the locations.
  \item The existing literature focuses on pairwise sequence alignment, which results in a high amount of information loss~\cite{medical,comparison,nergiz2008towards,gurung2014traffic,yarovoy2009anonymizing}.
  \item There is no unified metric to evaluate and compare the existing anonymization methods.
\end{itemize}

In this paper, we address the mentioned problems by proposing an enhanced anonymization framework termed machine learning based anonymization (MLA) to preserve the privacy of users in the publication of spatiotemporal trajectory datasets. MLA consists of two interworking algorithms: clustering and alignment. We have summarized our main contributions in the following bullet points.

\begin{itemize}
  \item By formulating the anonymization process as an optimization problem and finding an alternative representation of the system, we are able to apply machine clustering algorithms for clustering trajectories. We propose to use $k'$-means~\footnote{The prime notation on the top of variable ``$k$'' is to distinguish between the variable $k$ in the clustering algorithm and the variable $k$ used in the definition of $k$-anonymity.} algorithm for this purpose, as part of the MLA framework.
\item  We propose a variation of $k'$-means algorithm to preserve the privacy of users in the publication of overly sensitive spatiotemporal trajectory datasets.
  \item We enhance the performance of sequence alignment in clusters by considering multiple sequence alignment instead of pairwise sequence alignment.
  \item We propose a utility metric to evaluate and compare the anonymization frameworks.
\end{itemize}

MLA and all algorithms associated with it are applied on two real-life GPS datasets following different distributions in time and spatial domains. The experimental results indicate a significantly higher utility levels while maintaining $k$-anonymity of trajectories.

The rest of this paper is organized as follows. First, a comprehensive review of the currently existing literature is presented in Section \ref{Related work}, followed by the system model used in Section \ref{System model}. Next, the proposed framework is explained and analyzed in Sections \ref{approach} and \ref{Experiments}, respectively. Several real-world applications of the framework are elaborated in Section \ref{applications}, and finally, the paper is concluded in Section \ref{conclusion}.

%\IEEEraisesectionheading{\section{Introduction}\label{Introduction}}

\section{Related Work}\label{Related work}

Unfortunately, merely removing unique identifiers of users cannot protect their privacy, as databases can be linked to each other based on their quasi-identifiers. Doing so, adversaries can reveal sensitive information about the users and compromise their privacy. In this section, we review the existing approaches for the anonymization of spatiotemporal datasets.
%Most of the work in the area of privacy preservation for spatiotemporal trajectories is focused on achieving $k$-anonymity metric proposed in \cite{sweeney2002k}.

\subsection{Generalization Technique}

Generalization is currently one of the mainstream approaches for the anonymization of spatiotemporal trajectory datasets. The generalization technique is predicated on two interrelated mechanisms: clustering and alignment. Clustering aims at finding the best grouping of trajectories that minimizes a predefined cost function, and the alignment process aligns trajectories in each group.

The notion of k-anonymity was adopted in~\cite{nergiz2008towards} for anonymization of spatiotemporal datasets . The authors proved that the anonymization process is NP-hard and followed a heuristic approach to cluster the trajectories. The use of `edit distance' metric for anonymization of spatiotemporal datasets was proposed in~\cite{gurung2014traffic}. In this work, the authors target grouping the trajectories based on their similarity and choose a cluster head for each cluster to represent the cluster. Also, dummy trajectories were added to anonymize the datasets further. Yarovoy et al.~\cite{yarovoy2009anonymizing} proposed to use Hilbert indexing for clustering trajectories. The authors in~\cite{dong2018novel,liu2018location} chose to avoid alignment by selecting trajectories with the highest similarity as representatives of clusters. Poulis et al.~\cite{poulis2017anonymizing} investigated applying restriction on the amount of generalization that can be applied by proposing a user-defined utility metric. Takahashi et al.~\cite{takahashi2012cmoa} proposed an approach termed as CMAO to anonymize the real-time publication of spatiotemporal trajectories. The proposed idea is based on generalizing each queried location point with $k-1$ other queried location by other users, and hence, achieving $k$-anonmity.

The current state-of-art technique for applying gerelization to spatiotemporal datasets is based on generalization hierarchy (DGH) trees. In essence, DGH can be seen as a coding scheme to anonymize trajectories. We have categorized types of DGHs in the literature as:

\begin{itemize}
	\item Full-domain generalization: This technique emphasizes on the level that each value of an attribute is located in the generalization tree. If a value of an attribute is generalized to its parent node, all values of that attribute in the dataset must be generalized to the same level \cite{f1,f2,f3}.
	\item Subtree generalization: In this method, if a value of an attribute is generalized to its parent node, all other child nodes of that parent node need to be replaced with the parent node as well \cite{sub1,sub2}.
	\item Cell generalization: This generalization technique considers each cell in the table separately. One cell can be generalized to its parent node while other values of that attribute remain unchanged \cite{cell1,cell2,cell3}.
\end{itemize}

\subsection{Other Anonymization Techniques}

%In the second category, data custodian (server) plans to publish the spatiotemporal trajectories but needs to apply an anonymization technique, so that the identities of the trajectories are not revealed or linked to any other databases using quasi-identifiers. The authors in \cite{r6} adopted the notion of $k$-anonymity to trajectories and proposed an anonymization algorithm based on the generalization technique. They also explored the existing leaks of the privacy while releasing the spatiotemporal data. The work presented in \cite{r7} argued that there does not exist a fixed set of quasi-identifiers for all moving object databases and explained the pitfalls of simple adaptations used for achieving $k$-anonymity in spatiotemporal datasets. Rui Zhang et al. \cite{gurung2014traffic} presented an approach for clustering and anonymization of trajectories based on the 'edit distance' metric. Ding et al. \cite{ding2015trajectory} developed an algorithm called swap-and-match, which is based on changing the ID's of the trajectories once they reach an intersection.

Aside from the generalization technique, we have categorized the existing methods for the anonymization spatiotemporal datasets into three major groups:

\begin{itemize}
	\item \textbf{Perturbation} anonymizes location datasets by addition of noise to data;
	\item \textbf{ID swapping} swaps user IDs in road junctions to anonymize location datasets;
	\item \textbf{Splitting} divides trajectories into shorter lengths to anonymize location datasets.
\end{itemize}

The authors in~\cite{ding2015trajectory} proposed an algorithm that swaps the IDs of users in trajectories once they reach an intersection. Doing so, the algorithm prevents adversaries from identifying a particular user. Cicek et al.~\cite{cicek2014ensuring} made a distinction between sensitive and insensitive location nodes of trajectories. Their proposed algorithm only groups the paths around the sensitive nodes and exploits generalization to create supernodes.

Moreover, Cristina et al.~\cite{romero2018protecting} shifted the burden of privacy preservation in data publishing to the user side. The authors attempted to anonymize the data on the mobile phones before storage on the database as they would have more control over their privacy. Instead of clustering trajectories for anonymization, Cicek et al. in \cite{cicek2014ensuring} focused on the obfuscation of underlying map for sensitive locations. Brito et al. \cite{brito2015distributed} minimized the information loss during the data anonymization by suppressing key locations. The Local suppression and splitting techniques were considered for trajectory anonymization in~\cite{terrovitis2017local}. Although the proposed approach is useful for a predefined number of locations, it cannot be generalized to system models in which the users can make queries from an arbitrary location on the map. Naghizadeh et al.~\cite{naghizade2014protection} focused on the stop points along trajectories. A sensitivity measure is introduced in this work, which relies on the amount of time users spend in different locations. Sensitive locations are replaced or displaced with a less sensitive location to preserve the privacy of users. Jiang et al.~\cite{jiang2013publishing} considered the perturbation of locations by adding noise to preserve the privacy of users. Adding noise can generate fake trajectories that do not correspond to realistic scenarios.

\section{System Model}\label{System model}

%\subsection{Notation}\label{Notation}
We assume that a map has been discretized into an $\epsilon \times \epsilon$ grid and the time is discretized into bins with length $\epsilon_t$. Therefore, each point in the dataset represents a snapshot of a real-world location query including $x$-coordinate, $y$-coordinate, and time. The datasets with continuous time or space data can fit into our model using interpolation. The level of spatial-temporal granularity in discretization does not affect the effectiveness of the proposed model. In our model, we consider a spatiotemporal trajectory datasets denoted by $T$. The dataset consists of trajectories $tr_1,...,tr_n$ where $n$ represents the number of trajectories in the dataset ($T=\{ tr_1,...,tr_n\}\textrm{,} \ |T|=n$). The $i$-th trajectory $tr_i$ is an ordered set of $l_i$ spatiotemporal 3D points (i.e., $tr_i=\{ p_1,...,p_{l_i}\}\textrm{,} \ |tr_i|=l_i$). Each point $p_j$ is defined by a triplet $<x_j,y_j,t_j>$, where $x_j,y_j,t_j$ indicate the $x$-coordinate, $y$-coordinate, and the time of query, respectively.

%\subsection{Adversarial Model}\label{Adversarial Model}

\subsection{Generalization Model}\label{Hierarchical Tree Transformation}

Our proposed framework is based on the generalization technique to anonymize the spatiotemporal datasets. To apply this technique, we use the domain generalization hierarchy (DGH) trees and quantify the information loss accordingly.

\subsubsection{Domain Generalization Hierarchies}

DGH tree is defined formally in Definition \ref{def1}. To clarify the construction of DGHs, an example of such a tree for spatiotemporal datasets is provided in Example \ref{ex1}. In our model, we utilize three dimensions: x-coordinate, y-coordinate, and the time of queries in hours.

\begin{defn}\label{def1}
A DGH tree for an attribute $\mathcal{A}$, denoted as $H_{\mathcal{A}}$, is a partially ordered tree structure, which maps specific and generalized values of the attribute $\mathcal{A}$. The root of the tree is the most generalized value and is returned by the function $RT$.
\end{defn}
\begin{exmp}\label{ex1}
Consider an $4 \times 8$ map shown in Example \ref{ex1}. As can be seen in the figure, the generalization technique is applied by three DGH trees, each of them corresponding to one of the attributes. For instance, the $x$-coordinate attribute can have $8$ possible values ($0,1,...,7$). At the lowest level of the tree, each coordinate needs three bits of information to be shown that indicates the maximum information bits. As we go higher up the DGH tree, more information loss incurs, and less number of bits are used to represent the coordinates.
\end{exmp}

Each node on a DGH tree can be generalized by moving up one or multiple levels of the DGH. The process of generalizing $\textrm{node}_i$ to one of its parent nodes $\textrm{node}_j$ is denoted using $\textrm{node}_i \rightarrow \textrm{node}_j $. A special case of generalization, in which the node is generalized to the root of the DGH, is referred to as suppression.

\begin{figure*}[t]
\centering
\includegraphics[scale=1.23]{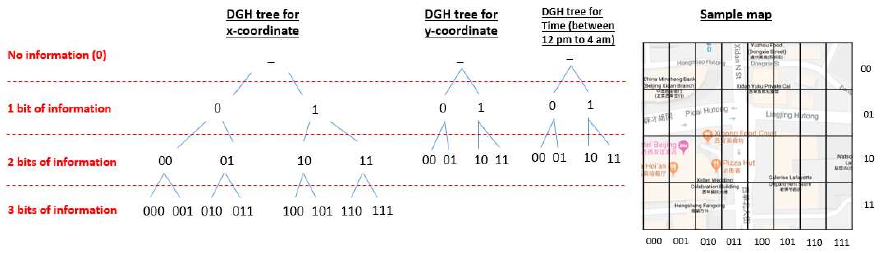}
\hspace{1em}
\centering
\caption{An example of DGHs for the attributes of spatiotemporal datasets.}
\label{tree}
\end{figure*}

For generalizing two nodes, it is necessary to find the lowest common ancestor (LCA). The LCA is a critical point in the generalization process due to its corresponding subtree that entails both the nodes and achieves the lowest information loss for the generalization of two nodes. The definition of LCA is given in Definition \ref{def3}.

\begin{defn}\label{def3}
The LCA of $\textrm{node}_i$ and $\textrm{node}_j$ in $H_{\mathcal{A}}$ is defined as the lowest common parent root of the two nodes. Function $LCA$ returns the LCA.
\end{defn}

For instance, in Example \ref{ex1}, if two leaf nodes `000' and `010' are to be generalized, their LCA corresponds to the parent node `0'. Hence, in the dataset, the x-coordinates `000' and `010' will be replaced by `0' to prevent adversaries from distinguishing between these two nodes.

\subsubsection{Information Loss}\label{Information Loss}

The information loss incurred by generalizing $\textrm{node}_i$ to $\textrm{node}_j$ in DGH $H_{\mathcal{A}}$ is defined as
\begin{equation}
LS(\textrm{node}_i,\textrm{node}_j) = \log_{2}{LF(\textrm{node}_j)}-\log_{2}{LF(\textrm{node}_i)}\textrm{ bits} ,
\end{equation}
where $LF(.)$ function returns the number of leaves in the subtree generated by a node, and $LS(.)$ function returns the loss incurred by the generalization of nodes. The calculation of information loss is elaborated in Example \ref{ex2}.

\begin{exmp}\label{ex2}
Consider the x-coordinate DGH tree given in Fig. \ref{tree}, the information loss incurred by generalizing node `$10$' to `$1$' can be calculated as $ \log_{2}{4} - \log_{2}{2}=1 \textrm{ bits}$.
\end{exmp}

Moreover, Lemma \ref{def33} can be used to derive the total loss incurred by the generalization of two nodes to their LCA.

\begin{thm}\label{def33}
The total loss incurred by generalizing $\textrm{node}_i$ and $\textrm{node}_j$ in $H_{\mathcal{A}}$ to their LCA, $\textrm{node}_p$, can be calculated as
\begin{align}
LS(\textrm{node}_i+\textrm{node}_j&,\textrm{node}_p ) =\nonumber \\ &LS(\textrm{node}_i,\textrm{node}_p) + LS(\textrm{node}_j,\textrm{node}_p).
\end{align}
\end{thm}
The total loss incurred during anonymization of a trajectory and a dataset are defined in Definitions \ref{def4} and \ref{def6}, respectively.
\begin{defn}\label{def4}
The total loss rendered by the generalization of trajectory $tr$ to achieve the anonymized trajectory $\overline{tr}$ with respect to attribute $\mathcal{A}$ can be calculated as
 \begin{align}
LS(\overline{tr},\mathcal{A} ) = \sum_{i=1}^{|\overline{tr}|}  LS(tr_i.\mathcal{A},\overline{tr}_i.\mathcal{A}).
\end{align}
where $tr_i.\mathcal{A}$ indicates the $i$-th location of the trajectory $tr$ with respect to the attribute $\mathcal{A}$. Here, $\mathcal{A}$ could denote $x$-coordinate, $y$-coordinate, or time.
\end{defn}

\begin{defn}\label{def6}
The total loss with respect to an attribute $\mathcal{A}$ in an anonymized dataset $\overline{T}$ can be computed as
\begin{align}
LS(\overline{T},\mathcal{A} ) = \sum_{\overline{tr} \in \overline{T}}^{|\overline{T}|}LS(\overline{tr},\mathcal{A} )
\end{align}
\end{defn}

%\begin{figure}[t]
%\centering
%\includegraphics[scale=.34]{Figures/architecture.eps}
%\hspace{1em}
%\centering
%\caption{An overview of the anonymization process for a given trajectory dataset.}
%\label{architecture}
%\end{figure}

\subsection{Privacy Model}

\subsubsection{Adversary Model}

In our work, we consider coordinates and the time of queries both to be quasi-identifiers, as they can be linked to other databases and compromise the privacy of users. We also assume that no uniquely identifiable information is released while publishing the dataset. However, the adversary may:

\begin{itemize}
  \item already know about part of the released trajectory for an individual and attempt to identify the rest of the trajectory. For instance, the adversary is aware of the workplace of an individual and attempts to identify his or her home address.
  \item already know the whole trajectory that an individual has traveled, but try to access other information released while publishing the dataset by identifying the user in the dataset. For instance, the published dataset may also include the type of services provided to users and if the adversary can identify a user by its trajectory, it can also know the services provided to that user.
\end{itemize}
To this end, our aim is to protect users against the adversary's attempt to access sensitive information that may compromise user privacy.

\subsubsection{Privacy Metric}

In this paper, we use a well-known metric called $k$-anonymity \cite{sweeney2002k} to ensure the privacy of users. The $k$-anonymity in our dataset implies that a given trajectory in the original dataset can at best be linked to $k-1$ other trajectories in the anonymized dataset. Definition \ref{def7} formally defines the $k$-anonymity in the context of dataset.
\begin{defn}\label{def7}
\textit{k-anonymous dataset:} A trajectory dataset $\overline{T}$ is a $k$-anonymization of a trajectory dataset $T$ if for every trajectory in the anonymized dataset $\overline{T}$, there are at least $k-1$ other trajectories with exactly the same set of points, and there is a one to one mapping relation between the trajectories in $\overline{T}$ and $T$.
\end{defn}

\subsubsection{Utility Metric}

K-anonymity metric ensures that the users are k-anonymous, and they can be identified from at least $k-1$ other users. Unfortunately, to achieve k-anonymity, significant loss of information can occur, which results in the much lower utility of published datasets. Moreover, as different anonymization techniques utilize various generalization schemes in the existing works, the information loss cannot be measured based on a unified metric as the one introduced in Section~\ref{Information Loss}. Therefore, it is necessary to develop a new metric to evaluate and compare the performance of anonymization schemes. In this work, we propose to use average released area per location to assess and compare various schemes. In the following, the calculation of average released area per location is explained.

Any anonymization approach aims to maximize the utility while preserving the privacy of users. Utility in generalization techniques refers to the area released for locations in the dataset. Consider a location in the dataset $T$ with coordinates $<x_1, y_1,t_1>$ and an arbitrary generalization function  $\mathcal{F}: T\rightarrow \overline{T}$. After the anonymization process, $<x_1, y_1,t_1>$ is generalized with a number of other locations $<x_2, y_2,t_2>$,..., $<x_a, y_a,t_a>$ in the dataset and an area $S$ would be released representing these locations. For instance, if generalization returns the minimum rectangle surrounding the locations. The generalized area is given by:

\begin{align} \label{l1}
S = (\underset{i}{\textrm{max}}\{ x_i\}-\underset{i}{\textrm{min}}\{ x_i\} )\times (\underset{i}{\textrm{max}}\{ y_i\} -\underset{i}{\textrm{min}}\{ y_i\} ).
\end{align}

Once the anonymization is conducted, assume that $n_1$ locations are generalized to area $S_1$, $n_2$ locations are generalized to area $S_2$,..., $n_b$ locations are generalized to area $S_b$. In this case, the average released area per location can be calculated as
\begin{align} \label{l2}
(\mathlarger{\sum}_{i=1}^{b} n_i\times S_i)/(\mathlarger{\sum}_{i=1}^{b} n_i),
\end{align}
in which no location belongs to more than one area. Average released area per location helps to understand how efficiently the data has been generalized and how much loss of utility has occurred by the generalization. Having $k$-anonymous locations, a smaller released area per location indicates a higher utility of data while preserving the privacy of users.

\subsection{Problem Formulation}\label{Architectural Overview}
The problem we seek to answer in this paper is formally presented in Problem 1 as follows.

\begin{problem}
    Given a trajectory dataset $T$, a privacy requirement $k$, quasi-identifiers $x$-coordinate, $y$-coordinate, and time, how to generate an anonymized dataset $\overline{T}$ which achieves the $k$-anonymity privacy metric and minimizes the total loss with respect to all quasi-identifiers, which can be explicitly formulated as
    \begin{align} \label{ee1}
    Minimize\{ LS(\overline{T},x) + LS(\overline{T},y)+ LS(\overline{T},t)\}.
    \end{align}
\end{problem}

\section{MLA}\label{approach}
In this section, we present our proposed framework, MLA, for anonymization of spatiotemporal datasets.

\subsection{Overview of the MLA Framework}
%remove this section

\begin{figure}[t]
\centering
\includegraphics[scale=1]{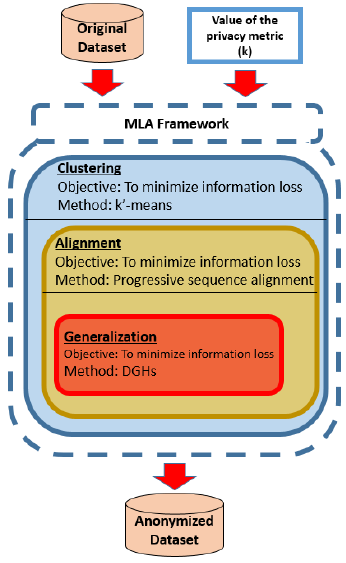}
\hspace{1em}
\centering
\caption{Overview of our proposed MLA framework.}
\label{architecture33}
\end{figure}

Fig.~\ref{architecture33} demonstrates the overview of our proposed framework. The original dataset and the value of $k$ are inputs of the framework, and the output is the anonymized dataset preserving the privacy of users. The MLA framework consists of three mechanisms working together to anonymize spatiotemporal datasets explained in the following:

\begin{itemize}
    \item \textbf{Generalization:} At the heart of MLA framework resides the generalization approach. The generalization process is conducted based on DGHs explained in Section \ref{Hierarchical Tree Transformation}.
    \item \textbf{Alignment:} For a given trajectory cluster, we propose to use progressive sequence alignment to find the arrangement of trajectories that results in the minimum information loss. Our approach for the alignment of trajectories is explained in Section~\ref{Alignment}.
    \item \textbf{Clustering:} At the highest level of the MLA framework, clustering is applied to seek for the most suitable grouping of trajectories that minimizes information loss. We propose to use $k'$-means clustering algorithm and a variation of it for overly sensitive datasets. Moreover, to have a baseline for comparison purposes, we develop a heuristic approach to cluster datasets. Our proposed clustering approaches are elaborated in Section~\ref{Clustering}.
\end{itemize}

\subsection{Alignment}\label{Alignment}

The process of alignment is defined as finding the best match between two trajectories in order to minimize the overall cost of generalization and suppression. The process of alignment between two trajectories has been studied in different domains mostly referred to as sequence alignment (SA). In this paper, we adopt a multiple SA technique called progressive SA \cite{chowdhury2017review} for anonymization of spatiotemporal trajectories.

\subsubsection{Progressive Sequence Alignment}\label{ProgressiveAlignment}
\begin{figure}[t]
\centering
\includegraphics[scale=.34]{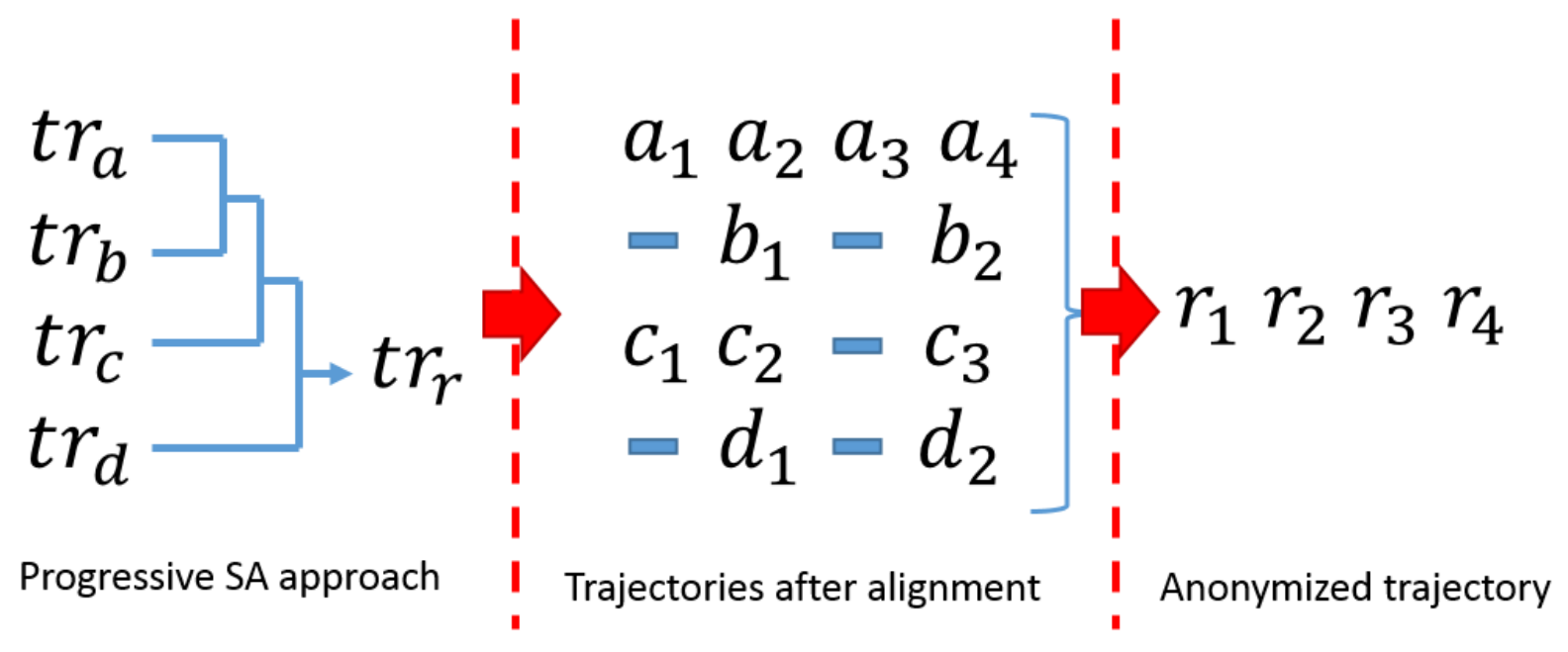}
\hspace{1em}
\centering
\caption{An overview of progressive SA for alignment of four trajectories and generating the anonymized trajectory.}
\label{example_figure1}
\end{figure}

The progressive SA is commonly used for SA of a set of protein sequences. Progressive SA is a greedy approach for multiple SA. As a part of the algorithm, pairwise alignment of the trajectories is required. We use dynamic SA for this purpose. Dynamic SA is based on dynamic programming and commonly used in DNA SA \cite{chen2017cmsa,le2017protein}. Fig. \ref{example_figure1} illustrates an example of how the progressive SA works for four hypothetical sequences $tr_a = \{a_1,\, a_2,\, a_3,\, a_4\}$, $tr_b = \{b_1,\, b_2\,\}$, $tr_c = \{c_1,\, c_2,\, c_3\}$ and $tr_d = \{d_1,\, d_2\}$ to generate the resultant aligned trajectory $tr_r = \{r_1,\, r_2,\, r_3,\, r_4\}$. The longest path $tr_a$ is chosen as the basis and it is aligned with a randomly chosen trajectory $tr_b$. The pairwise alignment process is implemented using dynamic SA. Then, the resultant trajectory is aligned with a third trajectory. The process continues until all trajectories are aligned. Instead of choosing the trajectories randomly during the progressive SA, the algorithm can choose the trajectory resulting in the lowest loss during the alignment. In Fig. \ref{example_figure1}, the way trajectory elements are located with respect to the longest path is referred to as the structure of the shorter path, and also, the spaces indicate the suppression operation during the alignment.

The dynamic SA algorithm is formally represented in Algorithm \ref{DynamicSA}. Dynamic SA is based on dividing the problem of finding the best SA to subproblems and storing the solutions of subproblems in a table or matrix referred to as $SAmatrix$ in the pseudocode. The objective is to achieve the minimal cost for SA. As before, the cost of alignment refers to the loss incurred during the alignment for different attributes of the sequence, which are $x$-coordinate, $y$-coordinate, and the time of the query.

%A subproblem generation for matching the first to $j$-th element of $tr_1$ ($tr_1=\{ p_1,\,,p_2,..., p_j \}$) with the first to $i$-th element of $tr_2$ ($tr_2=\{ q_1,\,,q_2,..., q_i \}$) can be given as 1) match $p_j$ and $q_i$; find the optimal alignment for $tr_1=\{ p_1,\,,p_2,..., p_{j-1} \}$ and $tr_2=\{ q_1,\,,q_2,..., q_{i-1} \}$ 2) suppress $p_j$; find the optimal alignment for $tr_1=\{ p_1,\,,p_2,..., p_{j-1} \}$ and $tr_2=\{ q_1,\,,q_2,..., q_{i} \}$ 3)  suppress $q_i$; find the optimal alignment for $tr_1=\{ p_1,\,,p_2,..., p_{j} \}$ and $tr_2=\{ q_1,\,,q_2,..., q_{i-1} \}$.

The algorithm starts by creating a $(m+1)\times (n+1)$ matrix ($SAmatrix$), where $m$ and $n$ denote the length of the trajectories. The matrix will be used to store the minimum cost of each cell of the grid. Moreover, a list called $code$ stores how cells have been reached. Cell $[j+1,i+1]$ can be reached from three cells $[j,i+1],\, [j+1,i],\, [j,i]$. Each path corresponds to one of the subproblems explained. After finding all values of the matrix and tracing back the list $code$, the outputs of the algorithm are the value of cell $[m,n]$ indicating the minimum value of the total loss ($TotLoss$) required for the dynamic SA, the aligned trajectory ($GenTraj$), and the structure of the shorter path compared to the longer path as $ShoTrajStr$.

\begin{algorithm}[t]
%\setstretch{1.3}
\DontPrintSemicolon % Some LaTeX compilers require you to use \dontprintsemicolon instead
\nonl \textbf{Required variables:} $tr_1=\{ p_1,\,,p_2,..., p_m \}$, $tr_2=\{ q_1,\,,q_2,..., q_n \}$, $H_x$, $H_y$, $H_t$\\

$SAmatrix \leftarrow \textrm{np.zeros}$($[m+1,n+1]$)\\
\For {$i\, \textrm{in}\, \textrm{range}(m)$} {
$Loss \leftarrow LS(p_i.x,\,rt(H_x))+ LS(p_i.y,\,rt(H_x))$\\
\nonl \quad \quad \quad $+ LS(p_i.t,\,rt(H_t))$\\
$SAmatrix[i+1,0] \leftarrow SAmatrix[i,0] + Loss$\\
}
\For {$i\, \textrm{in}\, \textrm{range}(n)$} {
$Loss \leftarrow LS(q_i.x,\,rt(H_x))+ LS(q_i.y,\,rt(H_x))$\\
\nonl \quad \quad \quad $+ LS(q_i.t,\,rt(H_t))$\\
$SAmatrix[0,i+1] \leftarrow SAmatrix[0,i] + Loss$\\
}

$options \leftarrow \textrm{np.zeros}(3)$\\
$code\leftarrow$ list()\\
\For {$i\, \textrm{in}\, \textrm{range}(m)$} {
     \For {$j\, \textrm{in}\, \textrm{range}(n)$} {
     $Loss\leftarrow $ loss incurred by generalizing $p_i$ and $q_j$\\
    $options[0]\leftarrow SAmatrix[i,j]+Loss $\\
    $Loss\leftarrow$ loss incurred by suppressing $q_j$\\
    $options[1]\leftarrow SAmatrix[i+1,j]+Loss $\\
    $Loss\leftarrow$ loss incurred by suppressing $p_i$\\
    $options[2]\leftarrow SAmatrix[i,j+1]+Loss $\\
    $BestOption \leftarrow \textrm{np.argmin}$ $(options)$\\
    $SAmatrix[i+1,j+1] \leftarrow options[BestOption]$\\
    $code$.append(index of option with minimum value)\\
     }
}

$TotLoss \leftarrow SAmatrix[m,n]$\\
$GenTraj \leftarrow $ trace back the $code$ to generate the aligned trajectory\\
$ShoTrajStr \leftarrow $ trace back the $code$ to find out structure of shorter trajectory while alignment\\

\textbf{Return} $GenTraj, ShoTrajStr, TotLoss$
\caption{DynamicSA($tr_1$, $tr_2$, $H_x$, $H_y$, $H_t$).}
\label{DynamicSA}
\end{algorithm}

\subsection{Clustering}\label{Clustering}
Clustering can be seen as a search for hidden patterns that may exist in datasets. In simple words, it refers to grouping data entries in disjointed clusters so that the members of each cluster are very similar to each other. Clustering techniques are applied in many application areas, such as data analysis and pattern recognition. In this subsection, first, we develop a heuristic approach for clustering the spatiotemporal datasets. Heuristic approaches are widely used in the literature, as the problem is found to NP-hard. The heuristic approach is used as a baseline for comparison in our work, and it is not part of the MLA framework. Next, we propose a technique that enables the $k'$-means algorithm for clustering the spatiotemporal trajectories and extend our approach for sensitive locations by developing a variation of $k'$-means algorithm that guarantees privacy requirements for all users.

\subsubsection{Heuristic Approach}\label{Heuristic Approach}
\begin{algorithm}[t]
%\setstretch{1.3}
\DontPrintSemicolon

$NumOfClus \leftarrow \lceil \dfrac{|T|}{k} \rceil $\\
$T \leftarrow OriginalDataset$ \\
Let $Clusters$ be a two-dimensional array storing the clusters and their corresponding trajectories
%$Clusters \leftarrow$ list()\\

\For {$c\, \textrm{in}\, \textrm{range}(0,NumOfClus)$} {
    Select a trajectory randomly from $T$ and append it to $cluster[c]$ while removing it from $T$\\
 %   $Traj1 \leftarrow$ first trajectory in $T$\\
%    $AlignedTraj \leftarrow Traj1$\\
%    $cluster[c]$.append$(NewMember)$\\
%    $T.\textrm{remove}(Traj1)$\\
    \For {$i\, \textrm{in}\, \textrm{range}(1,k)$} {
%        $LossMemory = \textrm{zeros}(|T|)$\\
%        $TrajMemory = \textrm{zeros}(|T|)$\\
        \For {$j\, \textrm{in}\, \textrm{range}(1,|T|)$} {
            Add the trajectory to the cluster $cluster[c]$\\
            Align based on DynamicSA and store the information loss\\
            Remove the trajectory from $cluster[c]$\\
        }
        Append the trajectory resulting in the minimum loss to $cluster[c]$ and remove it from the dataset\\
%        $NewMember \leftarrow$ The trajectory ID with minimum loss memory\\
%        $cluster[c]$.append$(NewMember)$\\
%        $T.\textrm{remove}(NewMember)$\\
%        $AlignedTraj \leftarrow$ update based on $NewMember$

    }

}
$(\overline{T},\, Loss) \leftarrow $GenerateAnonymizedDataset($cluster$, $OriginalDataset$)

\textbf{Return} $(\overline{T},\, Loss) $
%\caption{HeuristicClustering($OriginalDataset,\, k$, $H_x$, $H_y$, $H_t$).}
\caption{HeuristicClustering($OriginalDataset,\, k$).}
\label{HeuristicClustering}
\end{algorithm}

The heuristic approach for clustering spatiotemporal trajectory datasets is detailed in Algorithm \ref{HeuristicClustering} and its helper function in Algorithm \ref{GenerateAnonymizedDataset}. The intuition behind the heuristic algorithm is to form the clusters by sequentially adding the most suitable trajectory that minimizes the total loss incurred by generalization and suppression for $x$-coordinate, $y$-coordinate, and the time of query, given their DGHs $H_x$, $H_y$, $H_t$.

The algorithm starts by calculating the number of clusters that need to be generated and making a duplicate of the dataset called $T$. Moreover, a two-dimensional list is created, which holds the trajectory IDs for each cluster. For each cluster (i.e., cluster $c$), the algorithm appends a random trajectory from $T$. This trajectory is removed for $T$ and would be the first member of the cluster $c$. Then, given the privacy requirement $k$, $k-1$ other members of the cluster are chosen in a greedy approach. For every remaining trajectory in the dataset, the algorithm calculates the information loss incurred by applying dynamic alignment and determine the trajectory that results in the minimum loss. The chosen trajectory will be added to the cluster and removed from the dataset. The process continues until all members of the cluster are chosen. After clustering the trajectories, the helper function GenerateAnonymizedDataset is called in order to generate the anonymized dataset ($\overline{T}$) and the total incurred loss.

The helper function (GenerateAnonymizedDataset) takes the original dataset and the two-dimensional list of clusters as inputs. The target of the algorithm is to find the total loss and anonymize the dataset. The algorithm starts by initializing the total loss to zero and creating an empty list ($\overline{T}$) to hold the generated anonymized dataset. Then, for each cluster, the progressive SA is applied to calculate the incurred loss in addition to the generalized trajectory. In the next step, the total loss is accumulated, and the generalized trajectory is appended to the anonymized dataset $\overline{T}$. Eventually, the anonymized dataset and the overall information loss happened due to alignment are returned.

\begin{algorithm}[t]
%\setstretch{1.3}
\DontPrintSemicolon % Some LaTeX compilers require you to use \dontprintsemicolon instead
%\nonl \textbf{Required variables:} $tr_1=\{ p_1,\,,p_2,..., p_m \}$, $tr_2=\{ q_1,\,,q_2,..., q_n \}$, $H_x$, $H_y$, $H_t$\\
Let the $TotalLoss$ store the total loss incurred by applying progressive SA\\
Let $\overline{T}$ be an empty set that will store the anonymized dataset\\
\For {$i\, \textrm{in}\, \textrm{range}(0,len(cluster))$} {
    Apply progressive SA on trajectories in $cluster[i]$\\
    Add the incurred loss to $TotalLoss$\\
    Append the generated trajectory to $\overline{T}$

}

\textbf{Return} $(\overline{T},\, TotalLoss) $
\caption{GenerateAnonymizedDataset($cluster$, $OriginalDataset$).}
\label{GenerateAnonymizedDataset}
\end{algorithm}

\subsubsection{$k'$-means Clustering Approach}

$k'$-means algorithm \cite{macqueen1967some} is an attractive clustering algorithm currently used in many applications, especially in data analysis and pattern recognition \cite{pal2017genetic}. The main advantage of the $k'$-means algorithm is simplicity and fast execution.

\begin{figure*}[!t] %[!t]
\normalsize
%\hrulefill%makes a line
\begin{align} \label{n3}
&\textrm{Total loss} = \underbrace{\mathlarger{\sum}_{i=1}^{|T|} (LS(tr_i.x,RT(H_{x}))+ LS(tr_i.y,RT(H_{y})) + LS(tr_i.t,RT(H_{t}))) }_\text{A}-\nonumber \\
&\quad \quad \quad \quad \quad \quad \quad \quad \quad \underbrace{ (\mathlarger{\sum}_{i=1}^{|cluster|} \mathlarger{\sum}_{j=1}^{|cluster[i]|} (LS(h_j.x,RT(H_{x}))+LS(h_j.y,RT(H_{y}))+ LS(h_j.t,RT(H_{t})) ))}_\text{B}.
\end{align}
\hrulefill
\end{figure*}

The algorithm aims to partition the input dataset into $k'$ clusters. The only inputs to the algorithm are the number of clusters $k'$ and the dataset. Clusters are represented by adaptively-changing cluster centres. The initial values of the cluster centres are chosen randomly. In each stage, the algorithm computes the Euclidean distance of data from the centroids and partition them based on the nearest centroid to each data. More formally, representing the set of all centroids by $C= \{c_1,\, c_,...,\, c_{k'} \}$, each point in the dataset, denoted by $x$, is assigned to a centroid that has the shortest Euclidean distance to the point. This can be written as
\begin{align} \label{e2}
\underset{c_i \in C}{\textrm{argmin}} \,dist(x,c_i)^2,
\end{align}
where the function $dist(.)$ returns the Euclidean distance between two points. Denoting the set of assigned data to the $i$-th cluster by $S_i$, new centroids are calculated in the second stage via
\begin{align}
c_i= \dfrac{1}{|S_i|} \mathlarger{\sum}_{x_i \in S_i} x_i.
\end{align}
The algorithm continues the same process until the values of centroids no longer change. The $k'$-means algorithm is guaranteed to converge \cite{fischer2018convergence}.

In the rest of this section, we first present a Lemma followed by explaining how the $k'$-means algorithm can be applied to trajectory datasets to reinforce the privacy preservation of users.

\begin{thm}\label{def8}
The total loss incurred by generalizing $\textrm{node}_i$ and $\textrm{node}_j$ with respect to $H_{\mathcal{A}}$ can be calculated as
\begin{align}
LS(\textrm{node}_i ,&\textrm{node}_j ) =\nonumber \\ &|LS(\textrm{node}_i,RT(H_{\mathcal{A}})) - LS(\textrm{node}_j,RT(H_{\mathcal{A}}))|.
\end{align}
\end{thm}

\begin{exmp}\label{ex22}

Lemma \ref{def8} provides an alternative way to calculate the information loss by generalizing $\textrm{node}_i$ and $\textrm{node}_j$ in a given DGH. For instance, based on Lemma \ref{def8}, the information loss incurred by generalizing node `$10$' to `$1$' in Fig. \ref{tree} (x-coordinate DGH),  can be calculated as $|(\log_{2}{8} - \log_{2}{2})-(\log_{2}{8} - \log_{2}{4})|
= 1$ bit.
\end{exmp}

Lemma \ref{def8} indicates that the loss incurred by generalizing two nodes is equal to the difference between losses incurred by their suppression. As before, for any clustering outcome of data, assume that $cluster$ is a two-dimensional list, in which the $j$-th element of the list returns the IDs of the trajectories in the $j$-th cluster. Moreover, we denote the $j$-th cluster head after generalization and suppression for all trajectories as $h_j$. Therefore, the total loss can be written as

\begin{align} \label{n1}
\textrm{Total loss} =  LS(\overline{T},x)& + LS(\overline{T},y)+ LS(\overline{T},t)\nonumber \\ =\mathlarger{\sum}_{j=0}^{k-1} &\mathlarger{\sum}_{tr \in cluster[j]}(LS(h_j.x,tr.x)\nonumber \\ &+LS(h_j.y,tr.y)+LS(h_j.t,tr.t)).
\end{align}
As explained in (\ref{ee1}), the objective of clustering algorithms is to minimize this equation. Therefore, using Lemma \ref{def8} the equation (\ref{n1}) can be written as
\begin{align} \label{n2}
&\textrm{Total }\textrm{loss} =\\ &\mathlarger{\sum}_{j=0}^{k-1} \mathlarger{\sum}_{tr \in cluster[j]}(|LS(h_j.x,RT(H_{x})) - LS(tr.x,RT(H_{x})|\nonumber \\ &\quad \quad+|LS(h_j.y,RT(H_{y})) - LS(tr.y,RT(H_{y})|\nonumber \\&\quad \quad+|LS(h_j.t,RT(H_{t})) - LS(tr.t,RT(H_{t}))| .
\end{align}
Rearranging (\ref{n2}), the objective equation can be found by minimizing total loss formulated in (\ref{n3}). This can be done by maximizing part B and minimizing part A. Since the cluster heads are generated based on the clustering algorithm, they cannot be used as part of the optimization process. Therefore, we aim at minimizing part A in (\ref{n3}).

Part A in the equation (\ref{n3}) refers to finding the total distance of each trajectory from DGH root of the attributes. Therefore, for each trajectory, a three-dimensional vector $<d_x,\, d_y,\, d_t>$ is constructed, where $d_x$, $d_y$, $d_t$ store the loss incurred by generalizing the $x$-coordinate, $y$-coordinate, and time, respectively. Having distances of all points from the roots, we cluster the trajectories using the $k'$-means algorithm. The algorithm clusters trajectories with a similar loss from the root in the same group. This process is particularly important as trajectory datasets usually include trajectories as short as one query to trajectories with hundreds of queries.

A major drawback of the $k'$-means algorithm is clustering the trajectories without any constraint on the minimum number of trajectories that needs to be in each cluster. Therefore, the algorithm might result in some of the clusters containing less than $k$ trajectories that violates the $k$-anonymity of trajectories. If the data is not extremely sensitive such as the data used in the military, it is usually acceptable to have a few trajectories below the $k$-anonymity criterion. As it will be demonstrated in Section \ref{Experiments}, the number of trajectories not achieving $k$-anonymity is close to or below $20\%$ of the trajectories based on the value of $k$ chosen for the privacy. To amend the naive $k'$-means algorithm for sensitive applications, we propose to use a variation of $k'$-means algorithm, which we call it iterative $k'$-means. The idea relies on running the $k'$-means algorithm iteratively to ensure that all clusters will achieve $k$-anonymity. Therefore, after each iteration of the $k'$-means algorithm, the clusters including at least $k$ trajectories are disbanded, and the trajectories are put back into the pool for the next iteration of the $k'$-means algorithm. This process continues until all clusters have at least $k$ members. Algorithm \ref{kmeans} represents the pseudocode of the iterative $k'$-means.

\begin{algorithm}[t]
%\setstretch{1.3}
\DontPrintSemicolon % Some LaTeX compilers require you to use \dontprintsemicolon instead
%\nonl \textbf{Required variables:} $tr_1=\{ p_1,\,,p_2,..., p_m \}$, $tr_2=\{ q_1,\,,q_2,..., q_n \}$, $H_x$, $H_y$, $H_t$\\

\While{\textrm{true}}{
run $k'$-means algorithm on dataset ($\# \textrm{clusters}=\lfloor \dfrac{\# \textrm{data trajectories}}{k}\rfloor $)\\
remove trajectories that belong to clusters with at least $k$ members from the dataset\\
\If{$\# len(\textrm{dataset})< 2*k$}{
cluster the remaining trajectories together\\

$\mathbf{break;}$}
}

%\textbf{Return} clusters
\caption{Pseudocode of iterative $k'$-means algorithm.}
\label{kmeans}
\end{algorithm}

\section{Experiments}\label{Experiments}

\begin{figure*}[t!]
    \centering
    \begin{subfigure}[t]{0.3\textwidth}
        \centering
        \includegraphics[scale=.65]{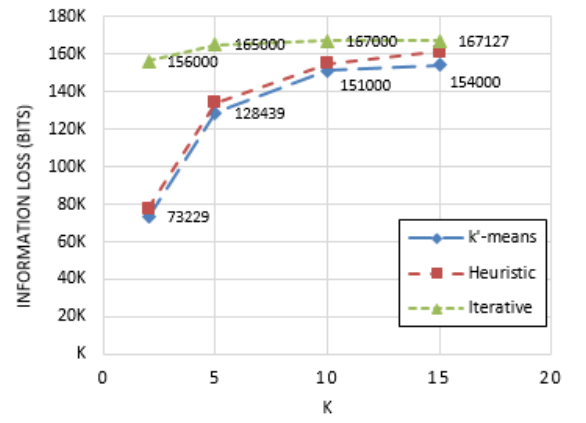}
        \caption{Total information loss}\label{a}
    \end{subfigure}%
    \hfill
    \begin{subfigure}[t]{0.3\textwidth}
        \centering
        \includegraphics[scale=.65]{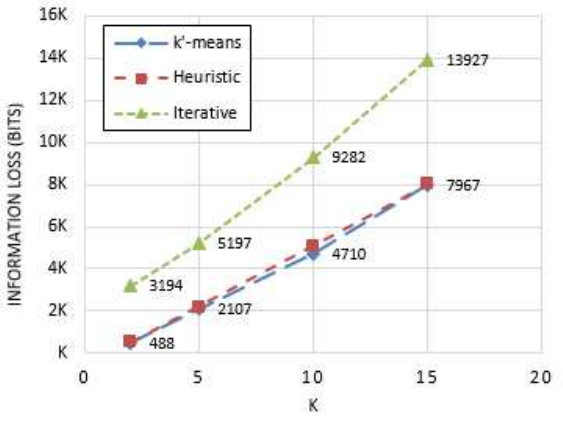}
        \caption{Average information loss per cluster}\label{b}
    \end{subfigure}
    \hfill
    \begin{subfigure}[t]{0.3\textwidth}
        \centering
        \includegraphics[scale=.65]{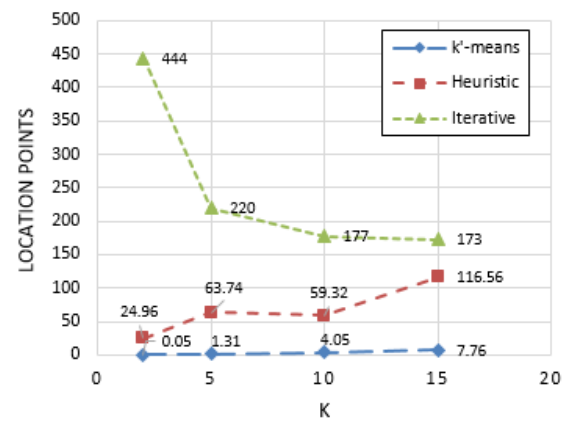}
        \caption{Average increase in length of trajectories}\label{c}
    \end{subfigure}
    \hfill
    \begin{subfigure}[t]{0.3\textwidth}
        \centering
        \includegraphics[scale=.65]{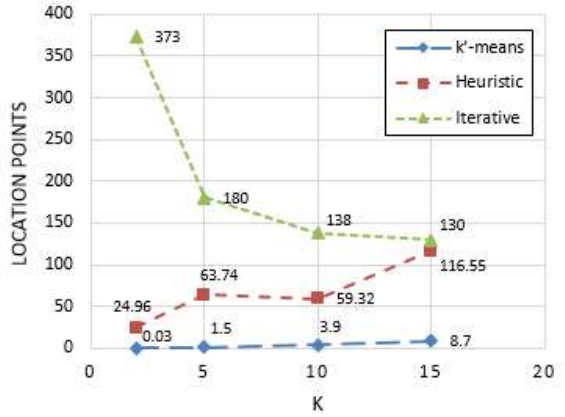}
        \caption{Average increase in length of trajectories per cluster}\label{d}
    \end{subfigure}%
    \hfill
    \begin{subfigure}[t]{0.3\textwidth}
        \centering
        \includegraphics[scale=.65]{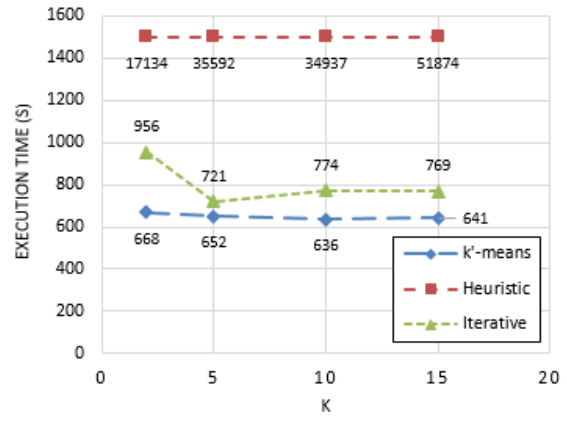}
        \caption{Total execution time (the heuristic algorithm's results are shown as a flat line with the values written below the line)}\label{e}
    \end{subfigure}
    \hfill
    \begin{subfigure}[t]{0.3\textwidth}
        \centering
        \includegraphics[scale=.65]{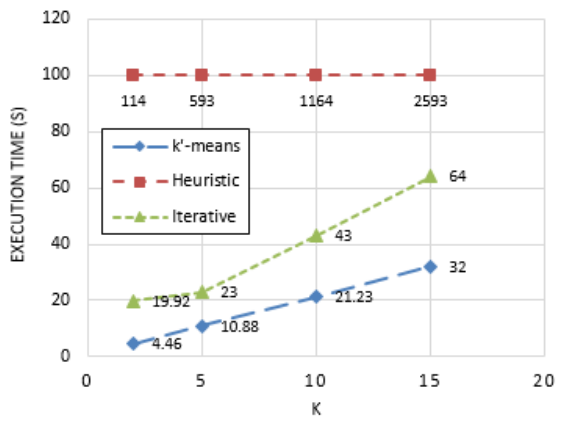}
        \caption{Average execution time per cluster (the heuristic algorithm's results are shown as a flat line with the values written below the line)}\label{f}
    \end{subfigure}
    \caption{Performance evaluation of MLA with different values of $k$.}
    \label{main1}
\end{figure*}

In our experiments, we use the data collected by Geolife project \cite{d1,d2,d3} in addition to the T-Drive dataset \cite{yuan2010t,yuan2011driving}, both published online by Microsoft. The Geolife and T-Drive datasets include the GPS trajectories of mobile users, and taxi drivers in Beijing (China), respectively. Each entry of the dataset is represented by the coordinates and the time of query. We have conducted our experiments on a $1\, km\times 1\, km$ central part of the Beijing map with the resolution of $0.01km\times 0.01km$ for each grid cell. The detailed statistics on the datasets are given in Table \ref{t1}. The various location privacy requirements ($k$) of the users are investigated for values $2$, $5$, $10$, and $15$. The experiments were performed on a PC with a $3.40$ GHz Core-i7 Intel processor, $64$-bit Windows $7$ operating system, and an $8.00$ GB of RAM. The Python programming language was used to implement the algorithms.

\begin{table}[th]
\caption{Statistics of datasets used in our experiments.}
\centering
\begin{tabular}{|>{\centering\arraybackslash}m{4cm} || >{\centering\arraybackslash}m{1.5cm} | >{\centering\arraybackslash}m{1.5cm} |}
 \hline
  Dataset& Geolife & TDrive   \\
 \hline Total number of samples & 47581 &  27916\\
 \hline Number of trajectories & 13561 &  301\\
 \hline Average number of samples per trajectory & 3.5 &  92.74\\
 \hline Average sampling time interval & 177 s &  1-5 s\\
 \hline Average sampling distance interval & 623 m &  5-10 m\\
 \hline
\end{tabular}
\label{t1}
\end{table}

\noindent We compare our work to prior methods as follows:

\begin{itemize}
	\item Many of prior approaches for the anonymization of spatiotemporal trajectory datasets use a greedy or so-called heuristic approach to anonymize datasets. In Section \ref{Heuristic Approach}, we explained and adopted this approach based on our system model. We use the heuristic approach in Section 5.1 as a baseline for comparison.
	\item The full comparison of the MLA framework with the recent work in \cite{comparison} is provided in Section \ref{Comparison 2}. The results are verified on both of the TDrive and Geolife datasets to ensure reliability.
	\item As MLA and the proposed algorithm in \cite{comparison} seek to fulfill different objectives, we have further evaluated the two frameworks based on random clustering. Doing so shifts the focus to the alignment of trajectories in each cluster.  The results are verified on both of the TDrive and Geolife datasets (Section 5.3).
	\item We also compare our alignment approach with the widely used static algorithm in \cite{medical} (Section 5.3).
\end{itemize}

\subsection{Performance Evaluation}

Fig. \ref{main1} presents the performance evaluation of MLA predicated on three clustering approaches developed in this paper. The algorithms have been investigated from three aspects: information loss, increase in trajectory length, and execution time. In all graphs, $x$-axis indicates $k$-anonymity requirement for the dataset. The total information loss and average information loss per cluster of algorithms are considered in Figs. \ref{a} and \ref{b}, respectively. Information loss, shown in the $y$-axis, indicates the total loss incurred while applying generalization and suppression on $x$-coordinate, $y$-coordinate, and the time of the query. The maximum possible incurred information loss for the whole dataset by suppressing all trajectories is $474572$ bits. This value is the upper bound on all anonymization algorithms. Note that this constant changes for different datasets. The main existing trend in Figs. \ref{a} and \ref{b} is that by increasing the value of $k$, the total incurred loss increases. This outcome meets our expectation as increasing the value of $k$ indicates having larger cluster sets, which results in the alignment of a higher number of trajectories in each cluster, and thereby, a higher total loss by the alignment. Among our proposed algorithms,  $k'$-means algorithm provides the best performance as it corresponds to minimum lost bits incurred by the generalization and suppression.

\begin{figure*}[t!]
    \centering
    \begin{subfigure}[t]{0.5\textwidth}
        \centering
        \includegraphics[scale=.75]{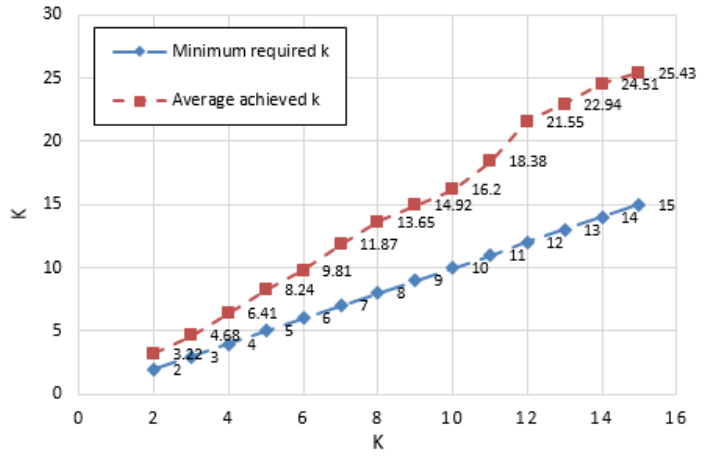}
        \caption{Average value of $k$ achieved by applying the $k'$-means algorithm}
    \end{subfigure}%
    ~
    \begin{subfigure}[t]{0.5\textwidth}
        \centering
        \includegraphics[scale=.75]{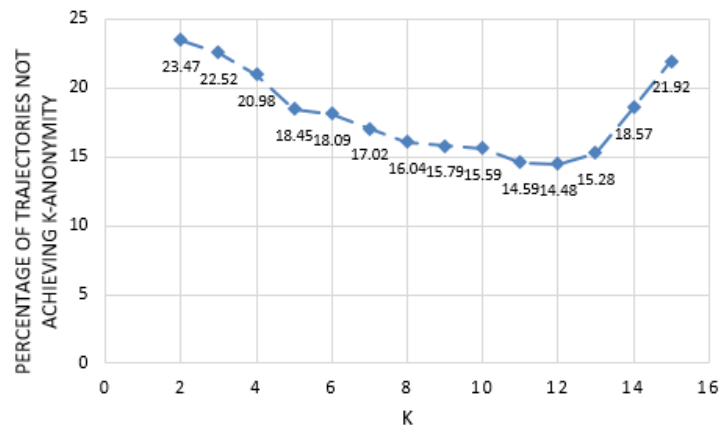}
        \caption{Percentage of users not satisfying $k$-anonymity requirement by applying the $k'$-means algorithm}
    \end{subfigure}
    \caption{Detailed performance evaluation of the $k'$-means algorithm.}
    \label{main_figure2}
\end{figure*}

The amount of information that $k'$-means algorithm preserves is higher than that of the heuristic approach, in which the most suitable trajectories are chosen to minimize the information. This trend can be seen for both of the total information loss of the dataset and the average information loss of dataset per cluster for different $k$ values. Such a trade-off exists, because some clusters contain a small number of trajectories not satisfying the $k$-anonymity requirement. The loss of privacy by $k'$-means algorithm is further analyzed in Fig. \ref{main_figure2} which will be explained later in this section. The iterative $k'$-means algorithm is constructed on top of the $k'$-means algorithm to ensure that all the trajectories satisfy the required privacy requirement. This is particularly important for sensitive applications, in which there are strict requirements for privacy preservation. The cost of having higher privacy for the iterative $k'$-means algorithm is a larger loss of information.

Figs. \ref{c} and \ref{d} present the average increase in the length of trajectories for the whole dataset and per cluster. Due to the alignment process, shorter trajectories may need to be aligned with longer trajectories, which result in an increase in the length of trajectories in the anonymized released dataset. The best performance among the algorithms is yielded by the $k'$-means algorithm with the lowest increase in the lengths of trajectories. Compared to other two approaches, the heuristic strategy performs better than the iterative $k'$-means with a smaller $k$, but as the $k$ value increases, the average increase in trajectory length converges due to large cluster size. Figs. \ref{e} and \ref{f} compare the total and average per cluster execution time of the different algorithms. Note that since the heuristic algorithm requires a significantly higher amount of time to run, it is shown on top of the graphs as a flat line with the corresponding values shown below it. The execution time of the $k'$-means and iterative $k'$-means algorithms are significantly lower than that of the heuristic algorithm and as expected the iterative $k'$-means consumes slightly more execution time as it has additional steps to ensure the $k$-anonymity of all trajectories.

\subsection{Detailed Analysis of $k'$-means Algorithm}

Overall, the detailed $k'$-means algorithm's results in satisfactory performance in terms of information loss, execution time, and the average increase in the length of trajectories. Moreover, the complexity of $k'$-means algorithm is of an order of the number of data entries for large datasets, whereas the order of the heuristic algorithm is proportional to the square of this number. Therefore, the $k'$-means algorithm has several significant advantages compared to the heuristic approach. Hence, if it is acceptable for the datasets to have a few trajectories below the $k$-anonymity requirement, then, it is more beneficial to use the $k'$-means algorithm instead of the heuristic or the iterative $k'$-means algorithm. This is usually true for datasets not entailing classified information. Therefore, we further analyze the performance of this algorithm in the remaining of this section and compare it to the state-of-art algorithms recently proposed. Also, note that in the rest of this paper when MLA is mentioned, the $k'$-means algorithms is adopted for clustering by default.

Fig. \ref{main_figure2} provides two graphs showing the details of the performance yielded by the $k'$-means algorithm. The first graph indicates the average value of $k$ achieved while applying the $k'$-means algorithm, and the second graph shows the percentage of trajectories that did not achieve the $k$-anonymity in the anonymization process with different values of $k$. In Fig. \ref{main_figure2}(a), it is evident that despite some of the trajectories losing their $k$-anonymity during the anonymization, the average value of anonymity achieved is above the minimum requirement. The value of the average gets even better as the value of $k$ increases. Fig. \ref{main_figure2}(b) shows the percentage of the trajectories not achieving the minimum required $k$-anonymity. This value is below $20\%$ on average, which means that over $80\%$ of the trajectories are guaranteed to at least have $k$-anonymity. The reason causing the uneven curves in the figure is because the number of clusters is divisible by $k$, which results in an additional cluster distorting the curves.

\subsection{Comparison}\label{Comparison 2}

We compare MLA with the static algorithm proposed in \cite{medical}, and recently published anonymization approach in \cite{comparison}. The idea behind the static alignment algorithm in \cite{medical} is that two trajectories are matched element by element without any shifts or spaces. In more details, the static algorithm attempts to match two sequences based on the same index. Therefore, each element of the first sequence $tr_1$ is aligned with an element having the same index in the other input trajectory $tr_2$. Based on our evaluation, the total incurred information loss is reduced by $7.2\%$ by using the proposed progressive SA algorithm. It must be noted that the dataset includes trajectories as large as hundreds of queries and as small as a single query from the location-based service provider. Therefore, matching these length-variant trajectories would impose a substantial information loss even for the best possible match of the sequences.

\begin{figure*}[t!]
    \centering
    \begin{subfigure}[t]{0.5\textwidth}
        \centering
        \includegraphics[scale=.75]{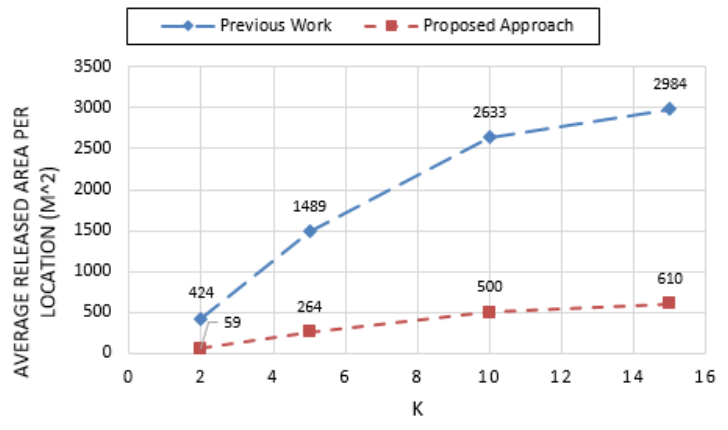}
        \caption{Geolife dataset}
    \end{subfigure}%
    ~
    \begin{subfigure}[t]{0.5\textwidth}
        \centering
        \includegraphics[scale=1.39]{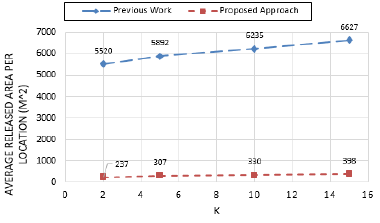}
        \caption{TDrive dataset}
    \end{subfigure}
    \caption{Comparison of MLA with the previous work proposed in \cite{comparison}.}
    \label{comp1}
\end{figure*}

\begin{figure*}[t!]
    \centering
    \begin{subfigure}[t]{0.5\textwidth}
        \centering
        \includegraphics[scale=.75]{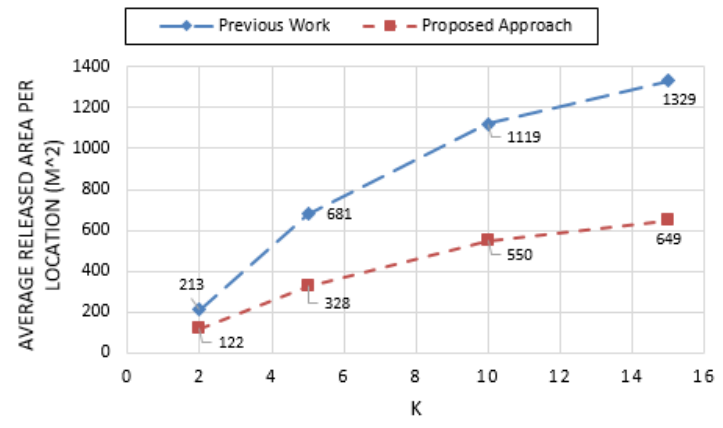}
        \caption{Geolife dataset}
    \end{subfigure}%
    ~
    \begin{subfigure}[t]{0.5\textwidth}
        \centering
        \includegraphics[scale=1.39]{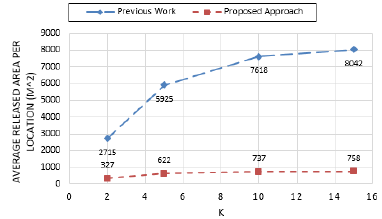}
        \caption{TDrive dataset}
    \end{subfigure}
    \caption{Comparison of MLA with the previous work proposed in \cite{comparison}, applying random clustering.}
    \label{comp2}
\end{figure*}

%\begin{figure}[t]
%\centering
%\includegraphics[scale=.69]{Figures/Previous_FreeStyle.eps}
%\hspace{1em}
%\centering
%\caption{Comparison of MLA with the previous work proposed in \cite{comparison}.}
%\label{comp1}
%\end{figure}

%\begin{figure}[t!]
%\centering
%\includegraphics[scale=.69]{Figures/Previous_Random.eps}
%\hspace{1em}
%\centering
%\caption{Comparison of MLA with the previous work proposed in \cite{comparison} when applying random clustering to both.}
%\label{comp2}
%\end{figure}

Fig. \ref{comp1} indicates the comparison result between our proposed anonymization technique and the recent generalization method proposed in \cite{comparison}. The authors in \cite{comparison} attempted to minimize the incurred loss of the anonymization by sorting out the spatiotemporal locations in the time domain and applying a heuristic approach for generalization. They also used a heuristic approach for clustering trajectories. Note that any anonymization approach aims to maximize utility while preserving the privacy of users. Utility in generalization techniques refers to the area released for locations in the dataset. Therefore, to have a fair comparison, we compare our work with the approach proposed in \cite{comparison} based on the average released area for locations. The metric is thoroughly explained in Section \ref{System model}. It can be seen from the figure that our proposed algorithm can significantly increase the utility of the generalization approach. In other words, the anonymized dataset has on average smaller released area per location while preserving the privacy of users. To further compare alignment approaches, in Fig. \ref{comp1}, we applied random clustering to group the trajectories, and then, used the alignment approach in our proposed work and the previous work to generate anonymized trajectories. As can be seen in the figure, our alignment approach outperforms the previous work by a higher utility of anonymized dataset.

\subsection{Discussion}\label{Discussion}

As can be seen in Fig. \ref{comp1}, the MLA framework has significantly improved the utility of data while achieving $k$-anonymity for the entries of datasets. A major reason for such an improvement is that MLA considers all three dimensions of time, $x$-coordinate, and $y$-coordinate together. Such consideration helps to minimize the overall cost and not just the utility in time or spatial domain. For instance, the Geolife datasets consists of sampling time interval of $177$ seconds with the average distance interval of $623$ meters, whereas the TDrive dataset has the average sampling interval of $1-5$ seconds and $5-10$ meters of sampling distance interval. Therefore, the two datasets enatail a highly different sparsity characteristic in time and spatial domain. However, as can be seen in Figs.~\ref{comp1} and \ref{comp2}, the MLA algorithm considers all three dimensions, and can significantly improve the utility in the process of anonymization.

In essence, the performance improvement in our proposed model is predicated on both the clustering and alignment of trajectories. In terms of the alignment, progressive SA has resulted in significant improvement of the alignment process. Such an impact can be seen in Fig. \ref{comp1}, where we apply random clustering, and therefore, the focus is on the alignment. As the figure suggests, utilizing a multiple SA technique such as progressive SA used in MLA provides major improvements to the utility of the anonymized datasets.

For clustering, as finding the optimal anonymization of the datasets is proven to be NP-hard, most of the literature has focused on following heuristic approaches to cluster the trajectories. We adopted such a heuristic approach for the system model of our paper and presented the results in Fig. \ref{main1}. Note that in the heuristic approach used in the figure, we are applying progressive SA alignment; therefore, the results show the improved version of the previously existing algorithms. As it was revealed in Fig. \ref{main1}, the $k'$-means algorithm can outperform such heuristic approaches in addition to having a much lower implementation complexity and processing time.

%As both mechanisms are interrelated, and both at minimizing the information loss, it would be hard to see the individual improvement achieved in each one separately.

%Also, a better performance of progressive SA compared to the static SA happens due to the dynamic programming used during the progressive SA which looks for the best match that minimizes the total loss.

%Overall, the detailed $k'$-means algorithm's results in satisfactory performance in terms of information loss, execution time, and the average increase in the length of trajectories. Moreover, the complexity of $k'$-means algorithm is of an order of the number of data entries for large datasets, whereas the order of the heuristic algorithm is proportional to the square of this number. Therefore, the $k'$-means algorithm has several significant advantages compared to the heuristic approach.

%Due to the alignment process, shorter trajectories may need to be aligned with longer trajectories, which result in an increase in the length of trajectories in the anonymized released dataset. The best performance among the algorithms is yielded by the $k'$-means algorithm with the lowest increase in the lengths of trajectories.

% It must be noted that the dataset includes trajectories as large as hundreds of queries and as small as a single query from the location-based service provider. Therefore, matching these length-variant trajectories would impose a substantial information loss even for the best possible match of the sequences.

\section{Applications}\label{applications}
In this section, we introduce several applications that we believe our work has the most impact on.

\subsection{Location-Based Data}

As the framework for anonymization presented in this paper considers location trajectories, one of the main applications of the framework is the privacy of location-based data. The use of location-based applications is more prevalent than any time before. Governments attempt to analyze the infrastructure using the location data and researchers use these data to investigate human behavior. Research has verified that even simple analytics on these published trajectory data would yield serious risk of users' privacy and even be capable of identifying users of location-based applications. \cite{sun2017asa}. Therefore, applying anonymization techniques such as the one we have developed in this paper is necessary to preserve the privacy of the users.

\subsection{Medical Records}

The recent advances in medical information technology have enabled the collection of a detailed description of patients and their medical status \cite{asan2018preferences}. Such data is usually stored in electronic medical record systems \cite{gates2018electronic,rahman2017systems,colleti2018evaluation}. Similar to spatiotemporal trajectories, many of the medical records need to be published by agencies and organizations. Unfortunately, research has shown that solely relying on de-identification is insufficient to protect users' privacy, as the medical records from multiple databases can be linked together to identify individual patients \cite{medical}. Therefore, there is an urgent need for viable algorithms to anonymize the medical data. The problem of anonymization in spatiotemporal trajectories is very similar to anonymization in longitudinal electronic medical records. This can be easily justified by the similar way, in which these data are stored. Assume a patient who has referred to medics several times in his or her lifetime. Each time the records of the patient are stored in a longitudinal dataset, in which the age and the diagnosed disease record are registered. These longitudinal records can be seen as a trajectory for the patient, and our proposed algorithms in this paper can be applied to anonymize a dataset of such longitudinal electronic medical records.

\subsection{Web Analytics}
Another important application of the framework developed in this paper is web analytics. Web analytics refers to analyzing online traces of users. Web analytics has become a competitive advantage for many companies due to the amount of detailed information that can be extracted from the data. Therefore, protecting the trajectories that the users explored on the Internet has become a major challenge for researchers. The similarity between spatiotemporal trajectories and web analytics can be well explained by the following example. For instance, Geoscience Australia is constantly recording and publishing the site logs users make on their website. The site log filename is composed of a four-digit station identifier, followed by a two-digit month and a two-digit year, e.g., ALIC0414 is the site log for the Alice Springs GNSS site that was updated in April 2014 \cite{aa}. Such a trajectory of logins to the website is analogous  to a spatiotemporal trajectory with three attributes. Therefore, the framework developed in this paper can be used to anonymize the online traces of users before publishing web browsing data.

\section{Conclusion }\label{conclusion}
In this paper, we have proposed a framework to preserve the privacy of users while publishing the spatiotemporal trajectories. The proposed approach is based on an efficient alignment technique termed as progressive sequence alignment in addition to a machine learning clustering approach that aims at minimizing the incurred loss in the anonymization process. We also devised a variation of  $k'$-means algorithm for guaranteeing the $k$-anonymity in overly sensitive datasets. The experimental results on real-life GPS datasets indicate the superior utility performance of our proposed framework compared with the previous works.

\bibliographystyle{IEEEtran}
\bibliography{MLA_Ref}

%\vskip -2\baselineskip plus -1fil
%\newpage

\begin{IEEEbiography}[{\includegraphics[width=1in,height=1.25in,clip,keepaspectratio]{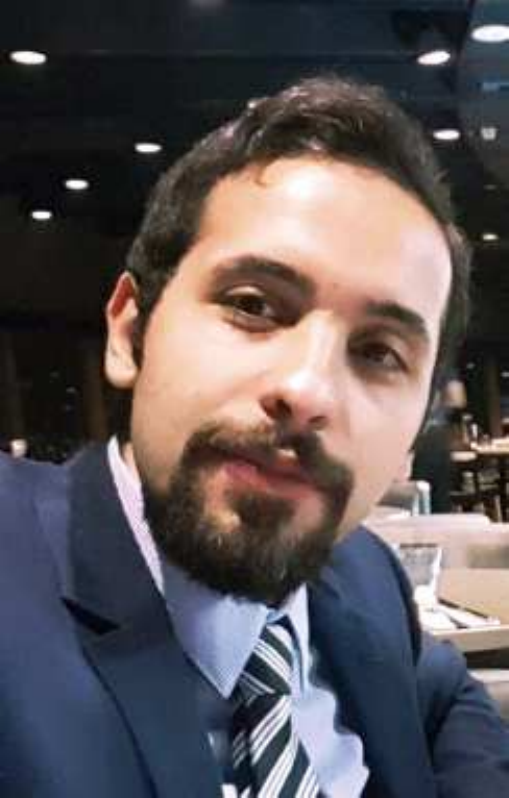}}]{Sina Shaham} received B.Eng (Hons) in Electrical and Electronic Engineering from the University of Manchester (with first class honors). He is currently an MPhil student at the University of Sydney. He has years of experience as a Data Scientist and Software Engineer in companies such as InDebted. His current research interests include applications of artificial intelligence in big data and privacy.
\end{IEEEbiography}

\vskip -2\baselineskip plus -1fil

\begin{IEEEbiography}[{\includegraphics[width=1in,height=1.25in,clip,keepaspectratio]{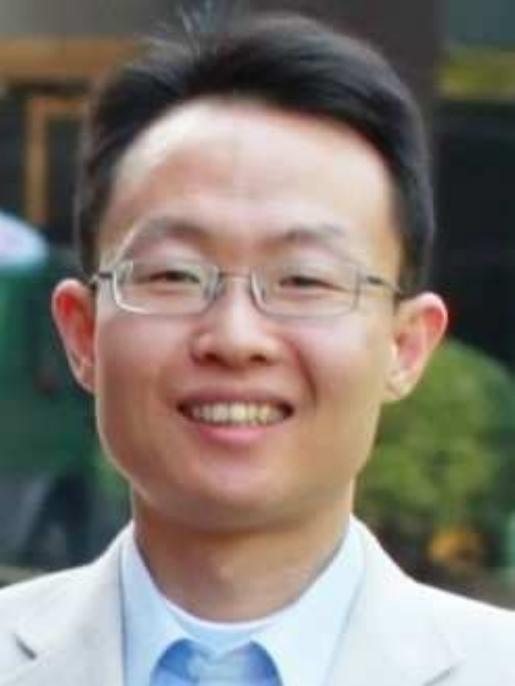}}]{Ming Ding} (M’12-SM’17) received the B.S. and M.S. degrees (with first class Hons.) in electronics engineering from Shanghai Jiao Tong University (SJTU), Shanghai, China, and the Doctor of Philosophy (Ph.D.) degree in signal and information processing from SJTU, in 2004, 2007, and 2011, respectively. From April 2007 to September 2014, he worked at Sharp Laboratories of China in Shanghai, China as a Researcher/Senior Researcher/Principal Researcher. He also served as the Algorithm Design Director and Programming Director for a system-level simulator of future telecommunication networks in Sharp Laboratories of China for more than 7 years. Currently, he is a senior research scientist at Data61, CSIRO, in Sydney, NSW, Australia. He has authored over 80 papers in IEEE journals and conferences, all in recognized venues, and about 20 3GPP standardization contributions, as well as a Springer book “Multi-point Cooperative Communication Systems: Theory and Applications”. Also, he holds 16 US patents and co-invented another 100+ patents on 4G/5G technologies in CN, JP, EU, etc. Currently, he is an editor of IEEE Transactions on Wireless Communications. Besides, he is or has been Guest Editor/Co-Chair/Co-Tutor/TPC member of several IEEE top-tier journals/conferences, e.g., the IEEE Journal on Selected Areas in Communications, the IEEE Communications Magazine, and the IEEE Globecom Workshops, etc. He was the lead speaker of the industrial presentation on unmanned aerial vehicles in IEEE Globecom 2017, which was awarded as the Most Attended Industry Program in the conference. Also, he was awarded in 2017 as the Exemplary Reviewer for IEEE Transactions on Wireless Communications.
\end{IEEEbiography}

\vskip -2\baselineskip plus -1fil

\begin{IEEEbiography}[{\includegraphics[width=1in,height=1.25in,clip,keepaspectratio]{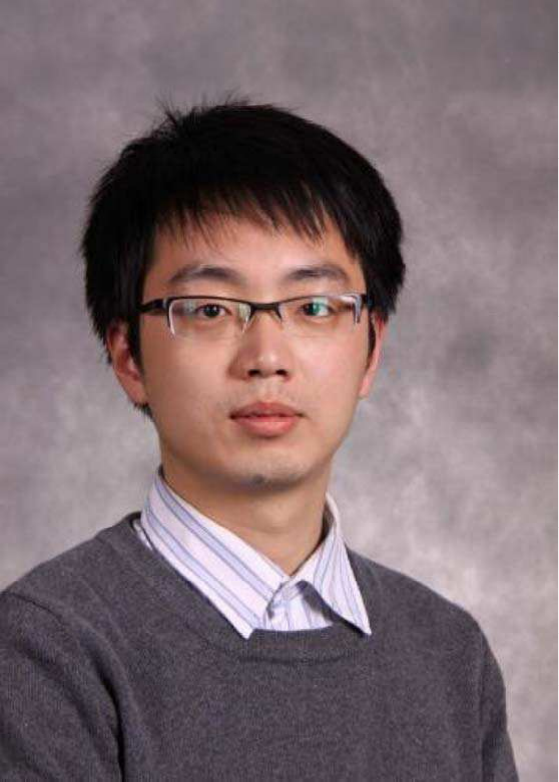}}]{Bo Liu} (M'10) received the B.Sc. degree from the Department of Computer Science and Technology from Nanjing University of Posts and Telecommunications, Nanjing, China, in 2004 and then he received the and MEng. and PhD degrees from the Department of Electronic Engineering, Shanghai Jiao Tong University, Shanghai, China, in 2007, and 2010 respectively. He was an assistant research professor at the Department of Electronic Engineering of Shanghai Jiao Tong University between 2010 and 2014, and a Postdoctoral Research Fellow at Deakin University, Australia, between November 2014 and September 2017. He is currently a lecturer in the Department of Engineering, La Trobe University, from October 2017. His research interests include wireless communications and networking, security and privacy issues in wireless networks.
\end{IEEEbiography}

\vskip -2\baselineskip plus -1fil

\begin{IEEEbiography}[{\includegraphics[width=1in,height=1.25in,clip,keepaspectratio]{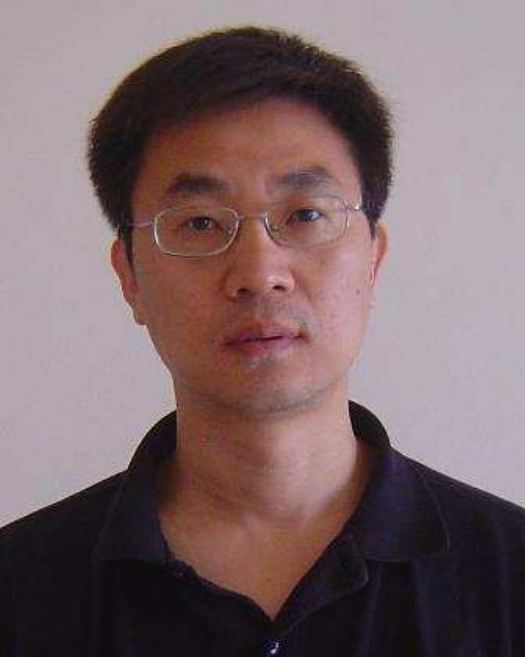}}]{Zihuai Lin} received the Ph.D. degree in Electrical Engineering from Chalmers University of Technology, Sweden, in 2006. Prior to this he has held positions at Ericsson Research, Stockholm, Sweden. Following Ph.D. graduation, he worked as a Research Associate Professor at Aalborg University, Denmark and currently at the School of Electrical and Information Engineering, the University of Sydney, Australia. His research interests include source/channel/network coding, coded modulation, MIMO, OFDMA, SC-FDMA, radio resource management, cooperative communications, small-cell networks, 5G cellular systems, etc.	
\end{IEEEbiography}

\vskip -2\baselineskip plus -1fil

\begin{IEEEbiography}[{\includegraphics[width=1in,height=1.25in,clip,keepaspectratio]{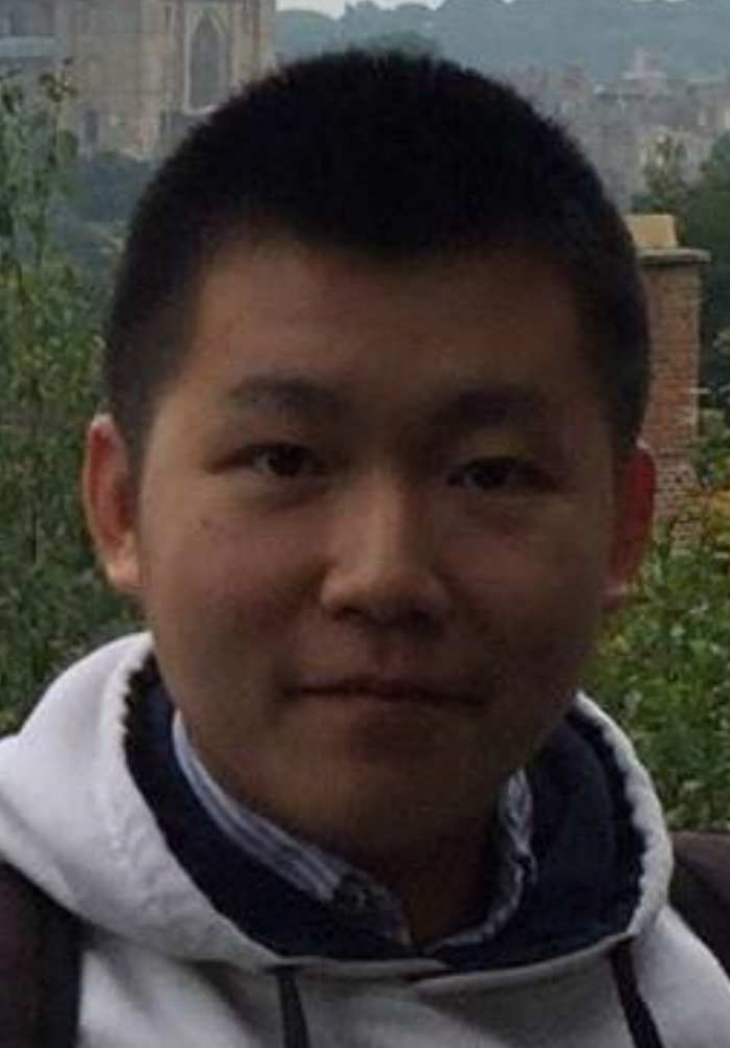}}]{Shuping Dang} (S'13--M'18) received B.Eng (Hons) in Electrical and Electronic Engineering from the University of Manchester (with first class honors) and B.Eng in Electrical Engineering and Automation from Beijing Jiaotong University in 2014 via a joint `2+2' dual-degree program. He also received D.Phil in Engineering Science from University of Oxford in 2018. Dr. Dang joined in the R\&D Center, Huanan Communication Co., Ltd. after graduating from University of Oxford and is currently working as a Postdoctoral Fellow with the Computer, Electrical and Mathematical Science and Engineering Division, King Abdullah University of Science and Technology (KAUST). He serves as a reviewer for a number of key journals in communications and information science, including {\scshape{IEEE Transactions on Wireless Communications}}, {\scshape{IEEE Transactions on Communications}} and {\scshape{IEEE Transactions on Vehicular Technology}}. His current research interests include artificial intelligence assisted communications, novel modulation schemes and cooperative communications.
\end{IEEEbiography}

\vskip -2\baselineskip plus -1fil

\begin{IEEEbiography}[{\includegraphics[width=1in,height=1.25in,clip,keepaspectratio]{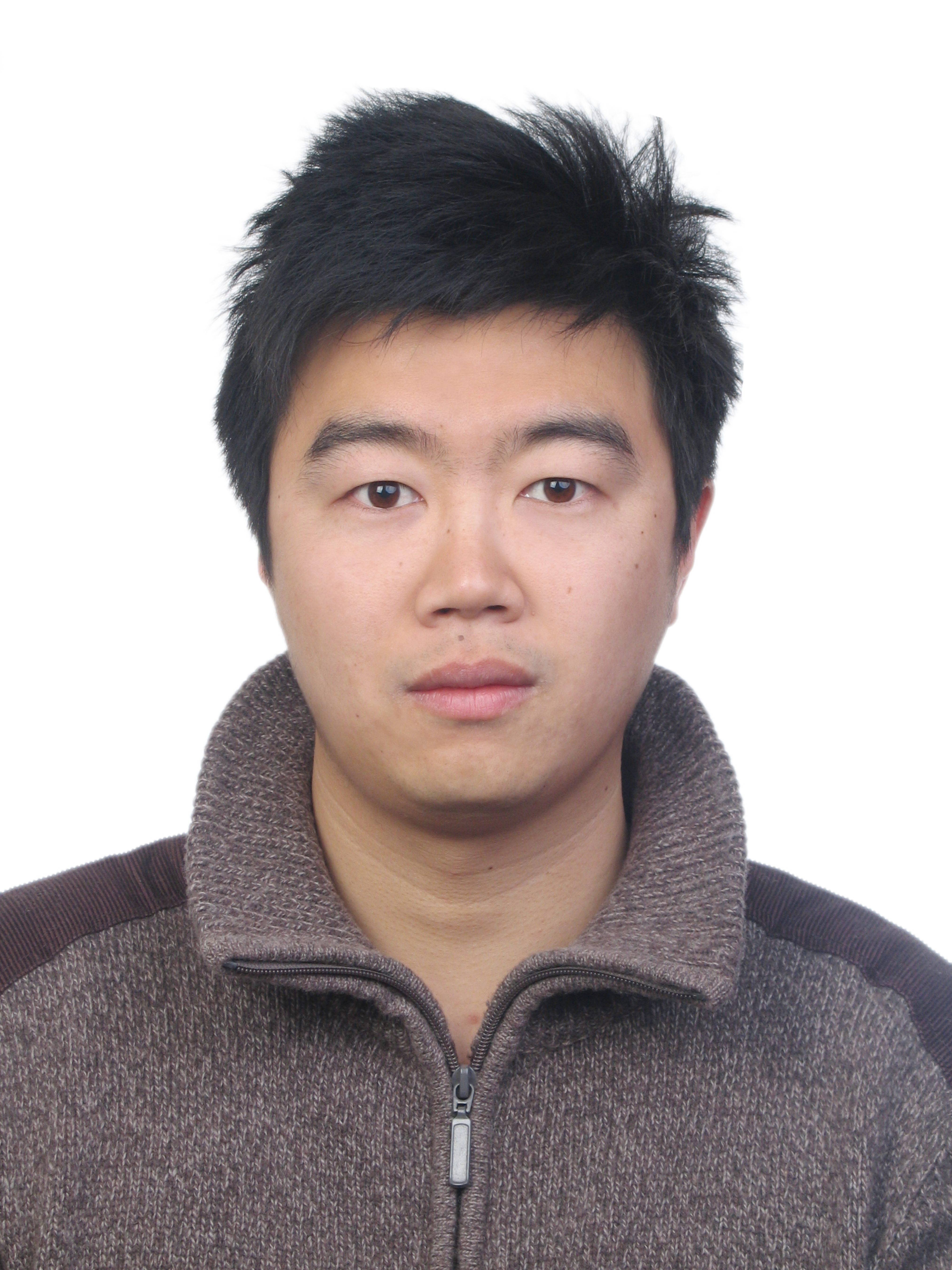}}]{Jun Li}  (M'09-SM'16) received Ph. D degree in Electronic Engineering from Shanghai Jiao Tong University, Shanghai, P. R. China in 2009. From January 2009 to June 2009, he worked in the Department of Research and Innovation, Alcatel Lucent Shanghai Bell as a Research Scientist. From June 2009 to April 2012, he was a Postdoctoral Fellow at the School of Electrical Engineering and Telecommunications, the University of New South Wales, Australia. From April 2012 to June 2015, he is a Research Fellow at the School of Electrical Engineering, the University of Sydney, Australia. From June 2015 to now, he is a Professor at the School of Electronic and Optical Engineering, Nanjing University of Science and Technology, Nanjing, China. His research interests include network information theory, ultra-dense wireless networks, and mobile edge computing.	
\end{IEEEbiography}

\end{document}